\pdfoutput=1

\documentclass[11pt]{article}

\usepackage[final]{acl}

\usepackage{times}
\usepackage{latexsym}

\usepackage[T1]{fontenc}

\usepackage[utf8]{inputenc}

\usepackage{microtype}

\usepackage{inconsolata}

\usepackage{graphicx}

\usepackage{booktabs}
\usepackage{ulem}
\usepackage{multirow}
\usepackage{lmodern}
\usepackage{pifont}
\usepackage{hyperref}
\newcommand{\checkbox}{\ding{113}} 

\title{Forecasting Future International Events: \\ A Reliable Dataset for Text-Based Event Modeling}

\author{
 \textbf{Daehoon Gwak$^{1}$}\thanks{~~Equal contribution.},
 \textbf{Junwoo Park$^{1,2}$$\textbf{}^{*}$},
 \textbf{Minho Park$^{1}$},
 \textbf{Chaehun Park$^{1}$},
 \textbf{Hyunchan Lee$^{3}$},\\
 \textbf{Edward Choi$^{1}$},
 \textbf{Jaegul Choo$^{1}$},
\\
\\
 \textsuperscript{1}KAIST AI \quad
 \textsuperscript{2}KRAFTON \quad
 \textsuperscript{3}Seoul National University,
\\
 \texttt{\{daehoon.gwak, junwoo.park, jchoo\}@kaist.ac.kr}
}

\begin{document}

\maketitle

\begin{abstract}
    Predicting future international events from textual information, such as news articles, has tremendous potential for applications in global policy, strategic decision-making, and geopolitics. However, existing datasets available for this task are often limited in quality, hindering the progress of related research. In this paper, we introduce WORLDREP (WORLD Relationship and Event Prediction), a novel dataset designed to address these limitations by leveraging the advanced reasoning capabilities of large-language models (LLMs). Our dataset features high-quality scoring labels generated through advanced prompt modeling and rigorously validated by domain experts in political science. We showcase the quality and utility of WORLDREP for real-world event prediction tasks, demonstrating its effectiveness through extensive experiments and analysis. Furthermore, we publicly release our dataset along with the full automation source code for data collection, labeling, and benchmarking, aiming to support and advance research in text-based event prediction.\footnote{~~https://github.com/eogns282/WORLDREP}
\end{abstract}

\section{Introduction}
Accurate prediction of future international events is essential for effective decision-making in international relations, global strategy, and security policy \citep{goldstein2011international, obrien2010crisis}. However, the dynamic nature of international relations, with its evolving conflicts and cooperation between multiple countries, presents a significant challenge for accurate prediction \citep{mellers2014psychological}. Recent global events, such as ongoing regional tensions and shifting economic dynamics among various countries, further complicate these challenges for policymakers and analysts. These complexities highlight the importance of developing reliable machine learning models capable of predicting changes in international relationships and the subsequent events they could trigger.

Existing approaches for text-based international events prediction like \citet{shi2024language} have heavily relied on datasets like the Global Database of Events, Language, and Tone (GDELT) \citep{leetaru13gdelt}  due to their comprehensive coverage and ease of use. By providing a comprehensive archive of global news and extracting relationships between subjects like countries, these datasets can be leveraged by machine learning models to tackle predictive tasks \citep{xing2018natural}. However, they have notable limitations:
\begin{enumerate}
\item \textbf{Overlooked Multilateral Relations}: GDELT frequently fails to capture complex interactions involving multiple countries, typically focusing on bilateral relationships. This simplification overlooks the intricate multilateral relations that are critical for understanding international dynamics. As analyzed in our study (detailed in Figure~\ref{fig:data_dist}(a) in Section~\ref{subsec:domainexperts}), this limitation becomes evident, highlighting the need for more comprehensive datasets.
\item \textbf{Inaccurate Labeling}: GDELT uses rule-based methods and basic machine learning techniques in some parts of the process \citep{saz2020empirical}, resulting in frequent mislabeling. Furthermore, the binary categorization into conflict or cooperation fails to capture the nuanced nature of international relations, lacking information to reflect the intensity of relationships and an `Unknown' category for indeterminate relationships. This oversimplification can lead to significant gaps in understanding and predicting complex international dynamics.
\end{enumerate}
Despite its significant contributions as a seminal work in the field, these limitations highlight the need for a more reliable and extensive dataset.

\begin{figure*}[h] 
    \centering
    \includegraphics[width=1.0\textwidth]{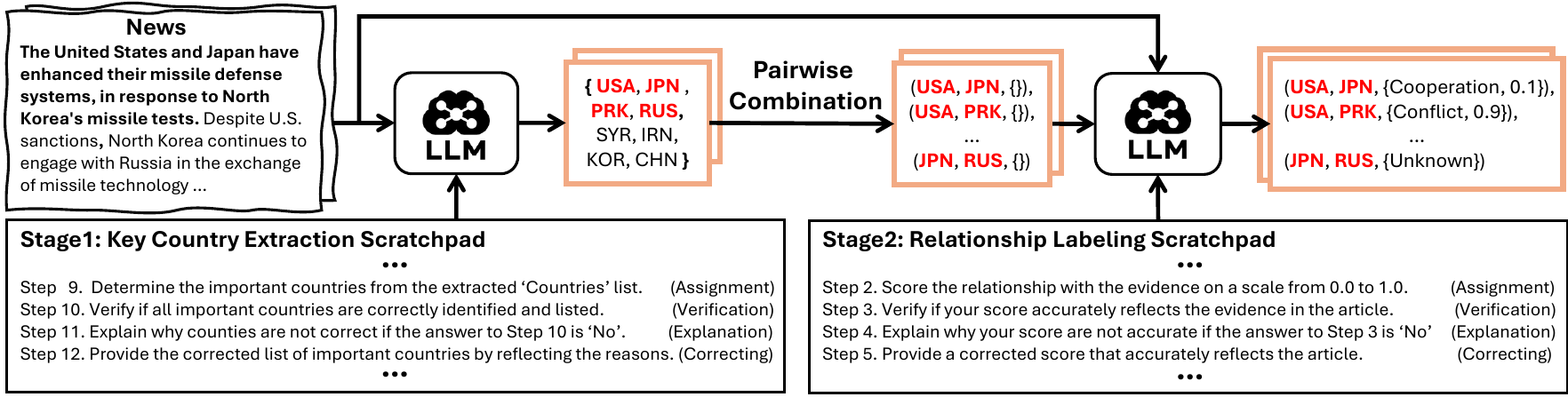} 
    \caption{A two-stage annotation process using LLMs to extract key countries and analyze their relationships from news articles. The first stage extracts countries mentioned in the news article, and in the second stage, relationships between these countries are labeled. Each stage employs a scratchpad including verification and correction steps to achieve efficient and accurate labeling.
    These scratchpads are summarized examples and the actual scratchpads and results of each stage can be found in Figure~\ref{main-fig:correcting-samples} and Appendix~\ref{app-sec:full-scratchpads}, respectively.}
    \label{fig:full_annotation_process}
\end{figure*}

In this paper, we introduce WORLDREP (WORLD Relationship and Event Prediction), a new dataset designed to overcome existing limitations by leveraging the advanced reasoning capabilities of large-language models (LLMs) \citep{brown2020language, radford2018improving}. To ensure reliable and efficient results, we developed a structured scratchpad \citep{nye2021show} with a self-correcting mechanism, incorporating a verification-correction pattern for annotations. This designed scratchpad allows for step-by-step reasoning, which helps in accurately identifying countries involved in multilateral events. The advanced contextual understanding of these models also ensures precise labeling of international relationships.

To validate the quality of WORLDREP, we conducted extensive experiments involving multiple steps. First, we obtained annotations from domain experts in international relations and political diplomacy. We then compared the number of extracted countries in our dataset to those identified by the experts, finding a high level of agreement. We also measured the alignment of our labels with the experts' labels, ensuring the consistency and reliability of our annotations. Next, we trained predictive models on WORLDREP and evaluated their performance using the expert-provided labels as the test set, achieving strong accuracy and F1 score. Leveraging these results confirming the reliability of our annotation process, we automated the labeling process and created a benchmark for predicting future international relations.

In summary, we present a comprehensive and reliable dataset that captures complex multilateral interactions. Using advanced LLMs, we accurately annotate relationship labels with a structured prompt design. We make both our dataset and the full automation source code publicly available to sustainably support further research and development in text-based event prediction and international relations. We believe this contribution will significantly advance the field by providing a solid foundation for future studies and applications.

\section{Dataset Construction}
Constructing a reliable dataset for research to predict future geopolitical events involves addressing several key challenges due to its complexity and ambiguity. We focus on two main enhancements compared to existing datasets such as GDELT: 1) capturing multilateral relations and 2) improving the accuracy of relationship labeling.

Our dataset construction process is designed to reflect the complexity of multilateral international relations more accurately. After collecting data, the process involves two main stages, Multi-Subject Extraction and Relationship Score Labeling, as illustrated in Figure~\ref{fig:full_annotation_process}. By following these steps, WORLDREP captures the intricate nature of global events and provides high-quality labels for research in predicting text-based international relations. The details of each step are explained below.

\subsection{Data Collection}
\begin{table}[]
  \centering
  \resizebox{\columnwidth}{!}{
  \begin{tabular}{lr}
    \toprule
    \textbf{Statistics} & \textbf{Value} \\
    \midrule
    
    Total Articles & 44,706  \\
    Total Unique Countries$^{\dagger}$ & 231 \\
    Avg. All Countries per Article & 4.45\\
    Avg. Key Countries per Article &  2.80 \\
    Avg. Labeled Pairs per Article & 3.31 \\
    Total Labeled Pairs & 147,931 \\
    Start Date & Feb 18, 2015 \\
    End Date  & May 29, 2024 \\
    
    \bottomrule
  \end{tabular}}
  \caption{Statistics of WORLDREP, including the number of articles, unique countries, and labeled pairs. The dataset covers the period from February 18, 2015, to May 29, 2024 with 231 countries. ${\dagger}$: Total number of codes is 249.}
  \label{tab:dataset-stats-table}
\end{table}
We collected a diverse set of news articles that could influence future international relationships between countries, resulting in a corpus of about 44,706 articles, including their dates and times. Following the approach taken by previous studies on future event prediction \citep{shi2024language}, we treat each news article as a single event with its corresponding occurrence time. Detailed data collection procedures are provided in the Appendix.

\subsection{Stage 1: Multi-Subject Extraction}
Identification of all relevant countries in international events is important for a comprehensive understanding of multilateral interactions. To extract multiple countries accurately, we incorporate a self-correcting mechanism within our LLM prompt design. As shown in Figure~\ref{fig:full_annotation_process}, this design consists of three tasks within a single prompt, allowing the LLM to extract, verify, and correct its output in single inference.
Specifically, our prompt instructs the LLM to perform the following tasks:

\vspace{1mm}

1. Extract important countries in the article.

\vspace{1mm}

2. Verify the accuracy of the extracted countries.

\vspace{1mm}

3. Provide corrections if necessary.
\vspace{1mm}

By doing so, we achieved significant improvement in the identification of all important countries compared to not using the last two tasks, verification and self-correction. The effects of verification and self-correction steps will be detailed in Section~\ref{sec:section3} comparing with extraction results by domain experts. The complete scratchpad is available in Appendix Figure~\ref{app-fig:full-country-scratchpads}.

\subsection{Stage 2: Relationship Score Labeling}

Accurate labeling of relationships is crucial for constructing a reliable dataset related to future geopolitical events. Given the inherent ambiguity and uncertainty in international relations, it is impractical to categorize interactions strictly as conflict or cooperation. Instead, we adopted a nuanced scoring system that reflects the complexities of these relationships. This scoring approach allows us to assign an `Unknown' category for cases where relationships cannot be clearly defined, providing a more accurate and comprehensive representation of international interactions.

Our scoring process includes the following steps:
\begin{enumerate}
\item \textbf{Pair Generation}: For each news article, identified subjects are paired to create possible relationships based on the article's content. For example, if four subjects are extracted in one news article with Stage 1, six pairs are generated.
\item \textbf{Scoring Relationships}: Each pair is evaluated and scored using a designed prompt that enables the LLM to:

1) Determine if there is evidence in the given article to predict a relationship between two countries.

2) If evidence is found, describe it and score the relationship on a scale from 0.0 (full cooperation) to 1.0 (full conflict). If no evidence is found, assign `Unknown' to the relationship.

3) For the pairs that have a score, verify whether the assigned score is biased or overly aggressive. Correct the score if necessary to ensure that it accurately reflects the relation.
\end{enumerate}

The full content of this designed prompt is illustrated in Appendix Figure~\ref{appfig:label_pad}. To increase the consistency and reduce the variance in the scoring process, we performed five separate scoring inferences for each relationship pair using the LLM. By averaging these five scores, we obtained a more stable and reliable relationship score. The statistics of our constructed dataset are detailed in Table~\ref{tab:dataset-stats-table}.

In the following section, we will validate the reliability of these labels through domain expert evaluations and various experiments. In addition, we will demonstrate the effectiveness of our prompt design, particularly the self-correcting mechanisms.

\section{Dataset Quality Evaluation}
\label{sec:section3}

To demonstrate the quality of WORLDREP, we conducted several experiments, including an evaluation against domain expert labels. Our evaluations in this section focused on the reliability of the labels and the validity of the designed prompting system.

\subsection{Domain Expert Annotation}
\label{subsec:domainexperts}
To establish a ground truth for our evaluation and to ensure the quality of our dataset, we engaged domain experts in political science and international relations. A group consisting of a professor and several graduates from a leading political science department, manually labeled the relationships in subset of 1,030 selected articles that have fully annotated samples in both GDELT and our dataset. These specific articles were chosen due to the limited number of fully annotated samples in GDELT, enabling a direct quality comparison between GDELT and WORLDREP. The experts labeled the relationships as either conflict, cooperation, or unknown if the relationship was indeterminable. The labeling process was supervised by the professor to ensure consistency, and detailed guidelines, example samples, and other relevant information about the labeling process are provided in Appendix. The domain expert labels for the selected articles are also made publicly available to facilitate future research.

\paragraph{Label Alignment with Domain Experts}
\begin{table}[t]
  \centering
  \scriptsize
  \resizebox{\columnwidth}{!}{
  \begin{tabular}{lcccc}
    \toprule
    \textbf{Dataset} & \textbf{Conflict } & \textbf{Cooperation } & \textbf{Unknown } & \textbf{Overall} \\
    \midrule
    \textbf{WORLDREP} & 84.8\% & 70.6\% & 73.7\% & 77.4\% \\
    \textbf{GDELT} & 48.8\% & 21.0\% & 0.0\% & 30.6\% \\
    \bottomrule
  \end{tabular}}
  \caption{Agreement of relationship labels between expert and our labels. Experts indicate an `Unknown' relationship between two countries due to insufficient information.}
  \label{tab:consistency}
\end{table}
\begin{table}[]
\centering
\resizebox{0.95\columnwidth}{!}{
    \begin{tabular}{l|c|c}
    \toprule
                              & \textbf{Avg. \# of Countries} & \textbf{F1 score}    \\ \midrule
    Initial Answer  & 4.06                     & 0.825 \\ \midrule
    Corrected Answer        & 5.60                    & 0.963 \\ \midrule
    Domain Experts         & 5.20                    & 1.000 \\
    \bottomrule
    
    \end{tabular}
}
\caption{Effectiveness of the self-correcting mechanism for extracting key countries from articles. The self-correcting mechanism improves the average number of identified countries, bringing it closer to the domain experts' average. In addition, it significantly increases the F1 score measured against domain expert annotations.}
\label{main-tab:self-correcting}
\end{table}

\begin{figure}[h]
    \vspace{-3mm}
    \centering
    \includegraphics[width=0.5\textwidth]{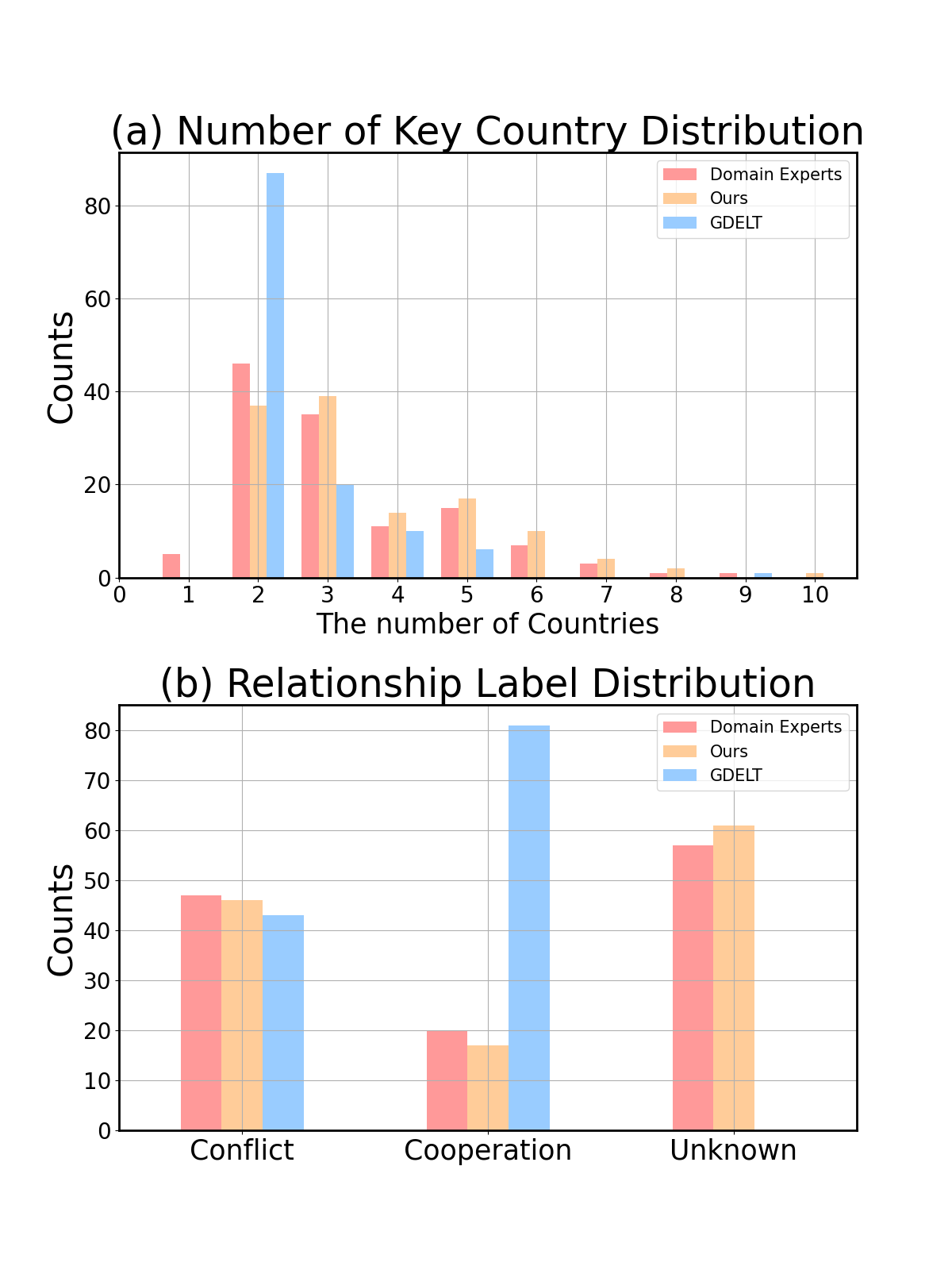} 
    \vspace{-14mm}
    \caption{(a) Distribution of the number of key countries. Our dataset aligns more closely with domain experts compared to GDELT, which tends to identify fewer key countries. (b) Distribution of relationship labels from different sources. The proportions of conflict, cooperation, and unknown labels in our dataset closely match those of domain experts, whereas GDELT lacks the unknown category, leading to significant imbalance.}
    \label{fig:data_dist}
\end{figure}

We compared the alignment of our labels with those from domain experts. To facilitate this comparison, we categorized our numerical scores into three classes: scores of 0.0-0.25 as cooperation, 0.75-1.0 as conflict, and 0.25-0.75 as unknown. Table~\ref{tab:consistency} shows the alignment rates for conflict, cooperation, and unknown categories, as well as the overall alignment rate.

As shown in Table~\ref{tab:consistency}, our dataset achieves alignment rates of 84.8\% for conflict, 70.6\% for cooperation, and 73.7\% for unknown, with an overall alignment rate of 77.4\%. In comparison, GDELT achieves 48.8\% for conflict, 21.0\% for cooperation, and does not include an "unknown" label, resulting in an overall alignment rate of 30.6\%.

These results demonstrate that WORLDREP not only has a higher overall alignment but also outperforms GDELT in both conflict and cooperation categories.  Additionally, Figure~\ref{fig:data_dist} (a) shows that the number of key countries identified by WORLDREP aligns more closely with domain experts compared to GDELT, which tends to identify fewer key countries. Figure~\ref{fig:data_dist} (b) illustrates that WORLDREP's label distribution more closely matches the domain experts' label distribution, further validating the accuracy and reliability of our dataset. The significant use of the "unknown" label by domain experts highlights the necessity of this category to accurately capture the complexity of international relationships.

\paragraph{Effectiveness of Self-correcting Mechanisms}
\begin{figure*}[h!]
    \centering
    \begin{minipage}{0.46\textwidth}
        \fbox{
            \begin{minipage}{\textwidth}
                \fontsize{10}{10}\selectfont
                \textbf{(a) Key Country Extraction} \\
                \rule{\textwidth}{0.4pt}
                \fontsize{7}{8}\selectfont
                \textbf{Article:} North Korea announced on Wednesday that it had launched its first reconnaissance satellite into orbit, ... (ellipsis) ... Additionally, it remains uncertain if Russia aided North Korea, suggesting that North Korea might have resolved technical issues ... North Korea experienced launch failures with rockets breaking into parts and falling into the Yellow Sea, East China Sea ...\\

                \textbf{Step~1:} Extract all countries mentioned in the article ... \\
                \textbf{Answer~1:} \underline{`RUS', `USA', `JPN', `CHN', `PRK', `KOR'} \\ 
                ...\\
                
                \textbf{Step 3:} Verify if the extracted `Countries' list ...\\
                \textbf{Answer 3:} \underline{No} \\
                
                \textbf{Step 4:} Explain why the `Countries' list is not accurate ...\\
                \textbf{Answer 4:} The provided list includes \uline{China (`CHN') which was mentioned only in the context of a sea name, not as an active country} in the article. \\
                
                \textbf{Step 5:} Provide the corrected list of countries ... \\
                \textbf{Answer 5:} \underline{`RUS', `USA', `JPN', `PRK', `KOR'} \\
                ... \\
                
                \textbf{Step 9:} From the extracted `Countries' list, determine the important countries based on the following criteria ...\\
                \textbf{Answer 9:} \underline{`RUS', `USA', `JPN', `PRK', `KOR'} \\
                
                \textbf{Step 10:} Verify if all important countries are correctly identified ... \\
                \textbf{Answer 10:} \underline{No} \\
                
                \textbf{Step 11:} Explain why the important countries are not accurately identified ... \\
                \textbf{Answer 11:} The list of important countries includes \uline{Russia (`RUS'), which, although mentioned in the context of providing potential assistance, is not directly related to the main event} or playing a significant role in the article. \\
                
                \textbf{Step 12:} Provide the corrected list of important countries ....\\
                \textbf{Answer 12:} \underline{`PRK', `KOR', `USA', `JPN'} \\
                \rule{\textwidth}{0.4pt}
                \fontsize{7}{8}\selectfont
                \textbf{Domain Expert Label}: {`PRK', `KOR', `USA', `JPN'}
            \end{minipage}
        }
    \end{minipage}
    \hspace{0.03\textwidth}
    \begin{minipage}{0.46\textwidth}
        \fbox{
            \begin{minipage}{\textwidth}
                \fontsize{10}{10}\selectfont
                \textbf{(b) Relationship Labeling} \\
                \rule{\textwidth}{0.4pt}
                \fontsize{7}{8}\selectfont
                 \textbf{Article:} Europe's slow vaccination program has given Russia an opportunity for diplomatic gain. A video conference between French President Macron, German Chancellor Merkel, and Russian President Putin discussed the potential registration and production of the Russian Sputnik V vaccine in the EU ...(ellipsis) \\

                \textbf{Countries:} {`FRA', `RUS'} \\

                \textbf{Step 1:} Determine if there is evidence to predict the relationship between `FRA' and `RUS' ... \\
                \textbf{Answer 1:} \underline{Yes} \\

                \textbf{Step 2:} Score the relationship based on the evidence ...\\
                \textbf{Answer 2:} \underline{Conflict with 0.7 score} \\

                \textbf{Step 3:} Explain your score with evidence from ...\\
                \textbf{Answer 3:} ... It highlights that the discussion between Macron and Putin over the vaccine has caused significant concern among many EU leaders and diplomats who view it as a tool being potentially used by Russia to exploit divisions within Europe. While there is some cooperative dialogue regarding the vaccine, ... \\

                \textbf{Step 4:} Verify if your score accurately reflects the full text of the article. ... \\
                \textbf{Answer 4:} \underline{No} \\

                \textbf{Step 5:} Explain why your score does not accurately reflect ...\\
                \textbf{Answer 5:} Although there is a significant element of distrust and concern about Russia's intentions, there are also instances of cooperation and willingness to engage from France's perspective. \uline{The initial score might have overemphasized the conflict aspects.} \\

                \textbf{Step 6:} Provide a corrected score ... \\
                \textbf{Answer 6:} \underline{Neutral with 0.5 score} \\

                \textbf{Step 7:} Explain why you gave the corrected score ... \\
                \textbf{Answer 7:} A score of 0.5 better reflects the nuanced nature of the relationship ... \\
                \rule{\textwidth}{0.4pt}
                \fontsize{7}{8}\selectfont
                \textbf{Domain Expert Label}: Neutral
            \end{minipage}
        }
    \end{minipage}
    \caption{Examples of self-correcting in our annotation process: (a) key country extraction and (b) relationship labeling. In (a), the initial list included `CHN' and `RUS' incorrectly. `CHN' was mentioned only as a sea name, and `RUS' was related to unclear facts. Through self-correcting including verification and explanation, they were removed. In (b), the initial relationship prediction followed the negative tone of the article. The self-correcting process adjusted it to a neutral score, aligning with domain expert opinions.}
    \label{main-fig:correcting-samples}
\end{figure*}
Our annotation process includes a self-correcting mechanism to enhance the accuracy and reliability of our labels. This mechanism allows the model to verify and correct its initial responses based on a comprehensive analysis of the article's content.

Figure~\ref{main-fig:correcting-samples} illustrates examples of our self-correcting process in both key country extraction and relationship labeling. In the key country extraction example (Figure~\ref{main-fig:correcting-samples}a), the initial list of countries extracted from the article is verified and corrected. Initially, the list included `CHN' (China) due to a sea name reference, which was later removed to provide a more accurate list of relevant countries. In the relationship labeling example (Figure 3b), the model initially scored the relationship between `FRA' (France) and `RUS' (Russia) as conflict with a score of 0.7. However, after verifying the context, the score was corrected to 0.5, reflecting a more balanced view of the cooperative and conflictual elements in the article.

Table~\ref{main-tab:self-correcting} quantifies the improvements achieved through self-correcting process. The average number of countries identified increased from 4.06 to 5.60, bringing it closer to domain experts' average of 5.20. Furthermore, the F1 score, which measures the alignment with domain expert labels, improved from 0.825 to 0.963 after correction, demonstrating a significant enhancement in label accuracy.

These results highlight the effectiveness of our self-correcting mechanism. By allowing the model to refine its initial outputs, we achieve a higher level of precision and reliability in our dataset. This approach not only captures a more comprehensive set of relevant countries, but also aligns more closely with the nuanced judgments of domain experts. 

Furthermore, we confirm that the latest LLM efficiently processes our scratchpads with numerous conditional instructions and format guidelines, resulting in few logical inconsistencies in its responses, as shown in Appendix Table~\ref{tab:logical-consistency}.

\subsection{Document Classification Experiments}
\begin{table}[]
  \centering
  \resizebox{\columnwidth}{!}{
  \begin{tabular}{ll|cc|cc}
        \toprule
        & & \multicolumn{2}{c}{\textbf{WORLDREP}} & \multicolumn{2}{c}{\textbf{GDELT}} \\
        \cmidrule(r){3-4} \cmidrule(r){5-6}
        & \textbf{Models} & \textbf{Accuracy} & \textbf{F1} & \textbf{Accuracy} & \textbf{F1} \\
        \midrule
        & \textbf{BERT} & 0.875 & 0.817 & 0.536 & 0.528 \\
        & \textbf{RoBERTa} & 0.859 & 0.803 & 0.480 & 0.476 \\
        & \textbf{ALBERT} & 0.823 & 0.719 & 0.331 & 0.310 \\
        & \textbf{XLNet} & 0.891 & 0.850 & 0.456 & 0.453 \\
        & \textbf{DistilBERT} & 0.823 & 0.736 & 0.540 & 0.531 \\
        & \textbf{ERNIE} & 0.865 & 0.823 & 0.608 & 0.591 \\
        \midrule
        & \textbf{Average} & 0.854 & 0.785 & 0.469 & 0.459 \\
        \bottomrule
      \end{tabular}
  }
  \caption{Comparison of test performance for various  models trained using different labeling schemes, ours and GDELT. The evaluation is based on domain expert-labeled test data, highlighting that models trained with our labeling scheme achieve significantly higher performance, indicating closer alignment with expert annotations.}
  \vspace{-3mm}
  \label{tab:model-performance-table}
\end{table}
The purpose of this subsection is to evaluate the reliability of our annotated labels by training various models on WORLDREP and comparing their performance against models trained on the GDELT dataset. Through these experiments, we aim to demonstrate the superiority of our labeling approach in providing consistent and accurate information about international relationship.

\subsubsection{Experimental Setup}
To evaluate the quality of our dataset, we designed a document classification task. The goal of this task is to classify the relationships between pairs of countries mentioned in news articles as either cooperative or conflictual. Specifically, each input consists of a document and two countries, and the task is to predict the nature of the relationship between these two countries based on the content of the news article.

The models were trained using labels from both WORLDREP and the GDELT dataset, but the evaluation of all models was performed using a test dataset labeled by domain experts. This approach ensures a consistent and unbiased comparison of the labeling quality across different datasets and allows us to benchmark our dataset against the highest standard of labeling quality.

To ensure a fair comparison, we focused on the samples where both our labels and the domain expert labels were not classified as "unknown." This setup allowed us to conduct a two-class classification task, making the comparison with GDELT more equitable. The models used in our evaluation include BERT \citep{devlin2018bert}, RoBERTa \citep{liu2019roberta}, ALBERT \citep{lan2019albert}, XLNet \citep{yang2019xlnet}, DistilBERT \citep{sanh2019distilbert}, and ERNIE \citep{Zhang2019ERNIEEL}. These pre-trained language models are widely recognized for their effectiveness in text classification tasks and are easy to fine-tune, making them ideal for evaluating the quality of our training labels. Details about the hyperparameters such as epochs, learning rates, and input instructions are provided in the Appendix.

\subsubsection{Experimental Results}

Using the document classification task, we evaluated the performance of various models trained on the same set of documents but with different labels: one set with our annotated labels and the other with GDELT. In addition, all models were evaluated using domain expert labels as ground truth.

As shown in Table \ref{tab:model-performance-table}, our labels consistently outperformed the GDELT labels in all models. Specifically, models trained on our labels achieved significantly higher accuracy and F1 scores compared to those trained on the GDELT labels. For example, the BERT model trained on our labels achieved an accuracy of 0.875 and an F1 score of 0.817, while the same model trained on GDELT labels achieved an accuracy of 0.536 and an F1 score of 0.528.

These results clearly demonstrate the superior quality of WORLDREP's labels. Models trained on our labels not only achieve higher overall performance, but also show consistent improvements across all evaluated models. Furthermore, the strong performance of models trained on our labels when evaluated with domain expert labels highlights the high alignment between our labels and those of the domain experts. This alignment is consistent with previous class distribution and consistency rate results, further validating the accuracy and reliability of our dataset. This highlights the effectiveness of our labeling approach in providing accurate and reliable relationship classifications, which are crucial for understanding international dynamics and predicting future events.

\section{Future Event Prediction}
\begin{table}[h]
  \small
  \centering
  \resizebox{0.85\columnwidth}{!}{%
    \begin{tabular}{lcc}
      \toprule
      \textbf{Model} & \textbf{Accuracy} & \textbf{F1 Score} \\
      \midrule
      GPT-4-Turbo & 0.615 & 0.615 \\
      GPT-4o & 0.612 & 0.610 \\
      Gemini-1.5-Pro & 0.612 & 0.611 \\
      Claude-3-Opus & 0.589 & 0.592 \\
      GPT-3.5-Turbo & 0.456 & 0.383 \\
      Llama3-8B & 0.487 & 0.471 \\
      Mistral-7B & 0.459 & 0.408 \\
      Phi3-mini-4k & 0.439 & 0.382 \\
      Llama2-13B & 0.453 & 0.360 \\
      \bottomrule
    \end{tabular}%
  }
  \caption{Accuracy and F1 scores for zero-shot future relation prediction across various LLM models. Models like GPT-4, Gemini-1.5, and Claude-3 demonstrate non-trivial performance in a challenging future prediction task with a three-class label distribution of 0.4:0.2:0.4.}
  \label{tab:future-event-results-table}
  \vspace{-3mm}
\end{table}

In this section, we provide a benchmark for text-based prediction of future relationships between countries. Predicting future international relations based on text data is a highly challenging task due to the complexities involved in modeling such events and the high cost of gathering quality-labeled datasets. In the previous section, we demonstrated the superiority of our labeling approach through comparisons with domain expert labels. Building on this foundation, we now leverage our automated labeling system to efficiently label new data, replacing the prohibitively expensive expert labeling process. This approach allows for flexible and timely updates of WORLDREP while maintaining high data quality. Based on the results of this labeling process, we provide a benchmark using the latest large models capable of performing this task. We hope that this benchmark will serve as a starting point for further modeling advancements in predicting future geopolitical events.

\begin{figure*}[h] 
    \centering
    \includegraphics[width=1.0\textwidth]{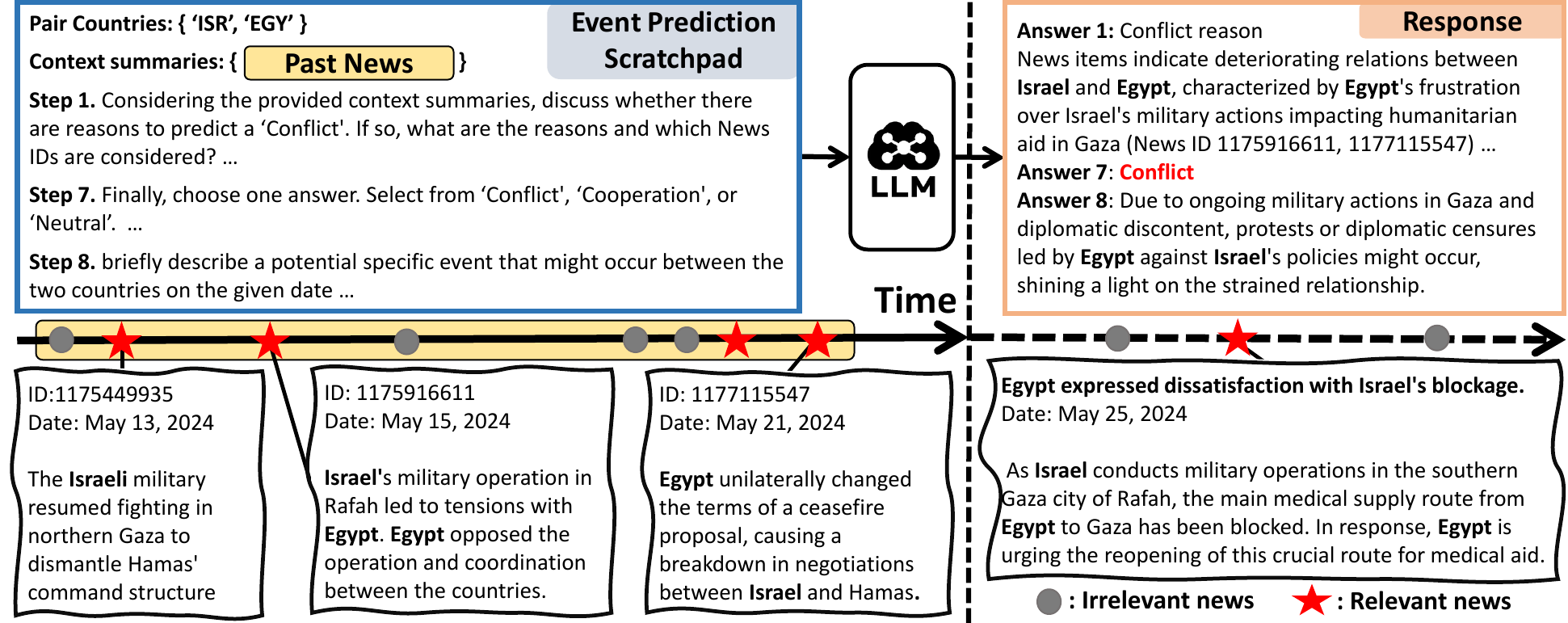} 
    \caption{An LLM-based prediction framework leveraging past news articles to forecast future relationships between countries. The figure shows the prediction process for the relationship between Israel and Egypt on May 25, 2024. Relevant past articles are highlighted, with summaries provided. The model predicts a conflict relationship (Answer 7), detailing reasons (Answer 1) and a potential specific event (Answer 8) based on the historical context and structured prompt.}
    \label{fig:future_pred_result}
    \vspace{-3mm}
\end{figure*}

\subsection{Experimental Settings}
\paragraph{Task Description}
The task involves predicting the relationship between two countries at a future date based on past news articles. The goal is to determine whether the relationship on the target date will be characterized by cooperation, conflict, or remain indeterminate. For example, consider the task of predicting the relationship between Country A and Country B on June 1, 2025. Given past articles, the task involves predicting whether the relationship on June 1, 2025, will be characterized by cooperation, conflict, or remain indeterminate. The prediction relies on the historical context provided by the past articles to assess the relationship between the two countries.

For the basic experiment, the model is provided with a fixed number of the most recent 15 articles up to the prediction date, under the assumption that recent events are more likely to influence future relationships. This context information provides an overview of the most recent interactions and events involving the countries. The articles are summarized to retain core information within the model's input limits. For this experiment, we used data from a two-week period, from May 15, 2024, to May 28, 2024, comprising a total of 353 articles.

The target setting specifies the relationship to be predicted between the two countries on the given future date. For this, we use the same categorization method described in the document classification section to convert our scoring system into categorical labels. This ensures consistency in the labeling process and allows for a clear definition of the relationship categories. We note that methods for news article selection and label categorization represent just a straightforward strategy, and more advanced techniques could certainly be employed in the future research.

\paragraph{Baseline Models} 
Successfully predicting future international relationships from text data requires the ability to understand and interpret context-rich complex information. Large Language Models (LLMs) are particularly suited for this task, as their ability to capture nuanced and complex contexts and generate insightful predictions has been extensively validated across various applications \cite{dhingra2022time, shi2024language}. Additionally, LLMs offer the potential to not only predict relationships but also generate detailed event descriptions, providing richer insights for future event prediction. Therefore, by leveraging both API-based and open-source LLMs, we aim to provide a comprehensive evaluation of their performance in this challenging predictive task. While LLMs are a focus of this benchmark, the benchmark experiments can be applied to other suitable models, allowing for broader exploration and evaluation in future event prediction tasks. The models include:
\begin{itemize}
\item \textbf{LLM API}: GPT-4-Turbo, GPT-4o \citep{achiam2023gpt}, Claude-3-Opus \citep{Anthropic2024Claude3}, Gemini-1.5-Pro \citep{team2023gemini}, GPT-3.5-Turbo \citep{brown2020language}
\item \textbf{Open-source LLM}: Llama3-8B \citep{Meta2024Llama3}, Llama2-13B \citep{touvron2023llama}, Phi3-mini-4k \citep{abdin2024phi}, Mistral-7B \citep{jiang2024mixtral}
\end{itemize}
These models are evaluated in a zero-shot setting, relying on their pre-trained knowledge and the provided context for predictions. The designed prompt and detailed model settings used for these evaluations are detailed in Appendix. Note that this experiment uses data in May 2024, which is beyond the knowledge cutoff dates of all these models.

\subsection{Experimental Results}
The performance of the models on the future event prediction task is summarized in Table~\ref{tab:future-event-results-table}. We report accuracy and macro F1 scores for each model. The range of accuracy across models is approximately 45-61\%, demonstrating that the models can make meaningful predictions even in a zero-shot setting. Additionally, API-based models generally exhibited better performance compared to open-source models. These results suggest that while the models show potential in predicting international event outcomes, there is significant room for improvement in their accuracy. These results suggest that the models are capable of predicting international event outcomes to a certain extent, but there is significant room for refining their accuracy. 
\paragraph{Qualitative Results}
To illustrate the practical applications of our prediction framework, we present a case study involving the relationship between Israel and Egypt. As shown in Figure~\ref{fig:future_pred_result}, the framework uses past news articles to predict the nature of the relationship on a future date. The context includes articles detailing recent interactions between the two countries, such as military operations and diplomatic tensions.

Using this context, the model predicts that the relationship between Israel and Egypt on May 25, 2024, will be characterized by conflict. The model's output includes not only the prediction but also a rationale for its decision. For example, the model identifies recent military actions and diplomatic tensions as key factors contributing to the conflict prediction.

This case demonstrates the model's ability to leverage historical context to make informed predictions about future international relationships, providing both the prediction and an explanation grounded in the provided news articles. Such detailed outputs underscore the potential of LLMs to assist in strategic decision-making and geopolitical analysis. Finally, for a detailed discussion on potential strategies to improve future relation prediction using LLMs, please refer to the Appendix.

\section{Conclusion}
In this paper, we introduced WORLDREP, a novel dataset designed to predict future international events from textual information, leveraging the advanced reasoning capabilities of large-language models (LLMs). WORLDREP addresses the limitations of existing datasets by capturing complex multilateral relations and providing high-quality labels validated by domain experts. Extensive experiments demonstrate the reliability and effectiveness of WORLDREP in real-world event prediction tasks. Furthermore, we established a benchmark for future event prediction, highlighting the potential and challenges of using LLMs for this purpose. We believe that WORLDREP and the automated labeling system will advance research in text-based event prediction and international relations, providing a solid foundation for future studies and applications.

\section*{Limitations} Despite the promising results, our work has several limitations. Firstly, while WORLDREP covers a broad range of international events, it may still miss less-reported but significant incidents. Secondly, the predictive models used in our experiments, although demonstrating meaningful performance, are not yet fully optimized for capturing the nuances of international relationships. This indicates potential for further improvement in model selection and fine-tuning. Lastly, our approach heavily depends on the capabilities of current LLMs, which may introduce biases inherent in their training data. Future work should aim to diversify data sources, enhance model accuracy, and address these biases through improved training methodologies.

\section*{Ethical Considerations} The development and application of predictive models in international relations come with ethical responsibilities. It is crucial to ensure that our models do not reinforce existing biases or contribute to misinformation. The data used in our models should be transparently sourced and carefully annotated to avoid misrepresentation. Furthermore, predictions made by our models should be interpreted with caution, recognizing the inherent uncertainties in forecasting international events. Policymakers and analysts using these models should complement the insights gained with human judgment and expertise to make informed decisions. We are committed to continually assessing and mitigating ethical risks associated with our research and its applications.

\section*{Acknowledgments}
This work was supported by Institute for Information \& communications Technology Promotion(IITP) grant funded by the Korea government(MSIT) (No.RS-2019-II190075 Artificial Intelligence Graduate School Program(KAIST)) and the National Research Foundation of Korea (NRF) grant funded by the Korea government (MSIT) (No. NRF-2022R1A2B5B02001913).

\bibliography{main}

\appendix
\section{Related Works}
\subsection{Benchmarks for International Event Prediction} Predicting international affairs involves a global scope and multiple intertwined interests, which makes it much more complex than forecasting individual social activities such as shopping, social media use, travel, or hospital visits, which are common in event prediction benchmarks~\cite{xue2023easytpp}. In international affairs, relying solely on the occurrence and types of past events is insufficient for accurate predictions. An accurate prediction must incorporate the contextual information surrounding international events. Consequently, existing studies~\cite{leetaru13gdelt, zou2022forecasting} have attempted to collect and process a vast amount of global news, treating news as an event in itself. Despite these efforts, the extensive range of collection and the need for expert knowledge in international affairs make relevant data more limited compared to social event datasets. Most international affairs prediction problems rely on the GDELT project\footnote{\url{https://www.gdeltproject.org/}}, which has created a huge news corpus by collecting news from around the world. However, the accuracy of mapping events to news through labeling is still lacking.

Recently, another type of dataset has been proposed for solving international affairs prediction problems using a QA-based approach, leveraging advancements in LLMs~\cite{zou2022forecasting, yan2023autocast, halawi2024approaching}. As LLMs have significantly improved the performance of natural language interfaces, QA-based approaches, and event QA datasets have been introduced. Notably, AutoCast~\cite{zou2022forecasting} includes queries about which events might occur at a future point, providing answers such as occurrence, numerical values, or selections from multiple choices. While the natural language interface allows for diverse questions, the predictions are often somewhat inaccurate, and the reliability of the answers is low, considering the difficulty of future prediction. Additionally, creating QA pairs is costly, resulting in fewer than $10K$ pairs, with even fewer questions related to international affairs.

\subsection{LLM-based Dataset Labeling and Generation}
In recent years, the use of LLMs for dataset generation and labeling has emerged as a significant trend in the ML community \citep{rel-wang2021want,yoo2021gpt3mix,liu-etal-2022-wanli,lee2023making,honovich2023unnatural,tan2024large-survey}.
Various studies show that LLMs can substantially reduce labeling costs and improve dataset quality. For instance, research by \citet{rel-wang2021want} suggests that GPT-3 can significantly lower labeling expenses. In another study, \citet{yoo2021gpt3mix} explore methods to enhance dataset diversity and richness through text augmentation. Additionally, \citet{liu-etal-2022-wanli} propose a collaborative approach between human workers and AI to create natural language inference datasets, demonstrating how such synergy can enhance dataset quality.

The utility of LLMs has also been proven in specific application domains. Research by \citet{goyal2022news} shows that summaries generated by GPT-3 receive higher evaluations from human annotators compared to those created by models fine-tuned on summarization datasets. Furthermore, \citet{wadhwa2023revisiting} confirm the effectiveness of LLMs in extraction tasks, highlighting their ability to accurately identify relationships between subjects. Lastly, \citet{lee2023making} address the use of LLMs to generate datasets for multi-choice question answering. These studies collectively demonstrate that LLMs are becoming innovative tools for dataset generation and labeling, opening new possibilities in the field of data science.

\subsection{Discussion about Improvement Strategies for Predicting Future Relations}
Based on the results of our experiments, we discuss potential strategies for improving future relation prediction using LLMs. We aim to understand how different factors influence model performance and identify ways to enhance predictive accuracy. Here, we outline two key strategies for improving the performance of LLMs in predicting future international relationships.

\textbf{Enhanced Context Retrieval}: Improving the methods for selecting context articles can significantly boost prediction accuracy. This includes leveraging more sophisticated retrieval algorithms and exploring more advanced embedding techniques. Advanced techniques to capture and represent context can ensure the consistency and relevance of context over extended prediction horizons, making retrieval more effective and accurate.

\textbf{Domain-Specific Fine-Tuning}: Fine-tuning LLMs on domain-specific datasets can help them incorporate more relevant knowledge, enhancing their predictive capabilities in specific contexts such as geopolitical events. For example, models might fail to make accurate predictions if the provided context does not include sufficient information. However, fine-tuning on a dataset specific to geopolitical events can improve the model's understanding and accuracy, even with minimal context. This suggests that fine-tuning can further refine the model's predictions by providing it with a more focused and relevant knowledge base.

These observations indicate that future research should focus on developing more advanced context retrieval techniques and fine-tuning strategies to improve model performance further. Enhanced context retrieval ensures that the most relevant information is used for predictions, while domain-specific fine-tuning can help models better understand and predict complex international relationships. By addressing these areas, the predictive capabilities of LLMs in forecasting future geopolitical events can be significantly improved.

\subsection{Future Applications of WORLDREP}
In this paper, we demonstrated that WORLDREP can serve as a reliable benchmark than existing datasets and can be effectively used to forecast text-based relationships between countries. Furthermore, we highlight that WORLDREP offers not only more accurate labels compared to other datasets but also provides significant value for traditional research areas, such as event prediction and time series forecasting.

Firstly, traditional event prediction models~\citep{pmlr-v119-zuo20a, chen2021neuralstpp}, including those based on Temporal Point Processes (TPP), have primarily relied on numerical or categorical data to predict next event by understanding frequency patterns or transitions between events. These models are typically benchmarked on datasets~\citep{xue2023easytpp} with simple event dependencies and dynamics. However, such datasets are insufficient for guaranteeing whether a model can learn the complex dynamics involved in applicable evnts in real-world such as international relations and economic events. In contrast, our dataset incorporates the rich semantic information embedded in news text, allowing for more sophisticated and reliable event prediction benchmark. By leveraging textual data, our model moves beyond simple predictions based on the numerical and categorical features, offering a new features for tackling more complex event prediction problems that require deeper insights into event dependencies.

Secondly, in time series forecasting~\citep{zhou2021informer}, our dataset enables models to go beyond simple periodic pattern recognition and account for the actual impact of international events on time series data, resulting in more accurate predictions. Events in international relations have a significant effect on economic uncertainty. This has a potential to improve time-series models. Previous studies~\cite{park23deep, wang2024timexer} have shown that much of the error in traditional time series forecasting models stems from unforeseen events. Also, even though AutoCon~\citep{park2024self} used to learn long-term correlation to achieve accurate forecasting, there are still rooms for improvement. Using our data set, we can validate time series models to see whether they can integrate semantic information from news events to improve prediction accuracy, not just learn periodic patterns. This enables a deeper analysis of how political and economic events influence time series patterns, providing a crucial foundation for reducing forecasting uncertainty.

\section{Dataset Details}
\label{sec:appendix}
\subsection{Domain Experts Annotation Process}
To accurately evaluate LLM-based annotations, reliable ground truth labels are essential. We employed human labels from international politics experts, and their involvement is crucial for obtaining trustworthy annotations from articles.

\begin{figure*}[h!]
    \centering
    \fbox{
        \begin{minipage}{1.0\textwidth}
        \fontsize{10}{12}\selectfont
    Based on the provided text \{News ID\}, please answer the following questions: \\
    
    The conflict between Country1 and Country2 began when Country2 police forcibly entered the Country1 embassy to arrest former Vice President Jorge Glas. Country1 considers this action a violation of the Vienna Convention on Diplomatic Relations and has filed a case with the International Court of Justice (ICJ), demanding Country2's expulsion from the United Nations. Conversely, Country2 argues that Country1's granting of asylum to Glas is a violation of international obligations and has filed a countersuit. Country2 President Daniel Noboa justified Glas's arrest, accusing the Country1 president of interfering in Country2 politics. This incident has garnered support for Country1 from several Latin American countries, leading to Country2's international isolation. \\

    1. Select all countries related to the content of the passage \{News ID\}.
 
    \begin{enumerate}
      \item  \checkbox\ Country1
      \item  \checkbox\ Country2
      \item  \checkbox\ Country3
      \item  \checkbox\ Country \# (These countries will be provided based on simple parsing rules.)
    \end{enumerate}
    
    2. Based on the content of the passage \{News ID\}, select the most appropriate description of the relationship between Country1 and Country2.
    \begin{enumerate}
      \item Conflict
      \item Cooperation
      \item Unknown
    \end{enumerate}
    
    3. Based on the content of the passage \{News ID\}, select the most appropriate description of the relationship between Country2 and Country3.
    \begin{enumerate}
      \item Conflict
      \item Cooperation
      \item Unknown
    \end{enumerate}

    ...
    \end{minipage}
    }
    \caption{Example of a questionnaire for domain experts for annotation requests.}
    \label{app-fig:userstudy}
\end{figure*}

We engaged three graduate students who majored in international politics under the supervision of a professor of international politics to label the relationships between major countries that appeared in articles. They worked under the guidance of a professor to ensure the quality of the labels. 

The process of identifying major countries went beyond simply checking if a country name appeared on a list; it involved understanding the context of the news articles to select key countries based on specific guidelines as follows:
\begin{enumerate}
\item Countries directly related to the major events or topics covered in the article.
\item Countries that are major actors or play specific roles in the article.
\end{enumerate}

Next, we extracted major countries from the article and then asked an expert to create possible country pairs from those countries and label the relationship between the two countries. The guidelines provided were as follows:
\begin{enumerate}
\item Choose between `Conflict' or `Cooperation' to describe the relationship between countries.
\item If it is impossible or ambiguous to choose based on a given article, `Unknown' can be selected.
\end{enumerate}
Using these guidelines, the experts carried out the annotations following the questions shown in Figure~\ref{app-fig:userstudy}. 

\subsection{Scratchpads Details}
\label{app-sec:full-scratchpads}

In our dataset construction process, we employed structured scratchpads to ensure the accurate extraction of important countries and the labeling of relationships in news articles. The scratchpads guide the model through a series of tasks to verify and correct its outputs, enhancing reliability.

Figures \ref{app-fig:full-country-scratchpads} and \ref{appfig:label_pad}
 illustrate the detailed prompts used for extracting important countries and relationship labeling, respectively.

The extraction scratchpad (Figure \ref{app-fig:full-country-scratchpads}) involves tasks such as extracting country codes, summarizing articles, and verifying accuracy. This process ensures that all relevant countries are identified and listed accurately.

\begin{figure*}[h!]
    \centering
    \fbox{
        \begin{minipage}{1.0\textwidth}
        \fontsize{10}{11}\selectfont
You are provided with an article. Your tasks are: \\

1. Extract all countries mentioned in the article as a comma-separated list of their 3-letter country codes (ISO 3166-1 alpha-3). If no countries are mentioned, return "None".\\

2. Summarize the article in up to 10 sentences, ensuring all mentioned countries are included.\\

3. Verify if the extracted `Countries' list includes all countries mentioned in the article. Answer `Yes' or `No'.\\

4. Explain why the `Countries' list is not accurate if the answer to step 3 is `No'. If the answer to step 3 is `Yes', return `None'.\\

5. Provide the corrected list of countries as a comma-separated list of 3-letter country codes (ISO 3166-1 alpha-3) if the answer to step 3 is `No'. If the answer to step 3 is `Yes', return `None'.\\

6. Verify if the `Summary' of the article content is accurate and includes all mentioned countries. Answer `Yes' or `No'.\\

7. Explain why the `Summary' is not accurate if the answer to step 6 is `No'. If the answer to step 6 is `Yes', return `None'.\\

8. Provide the corrected summary, ensuring it includes all mentioned countries and follows the 10-sentence limit if the answer to step 6 is `No'. If the answer to step 6 is `Yes', return `None'.\\

9. From the extracted `Countries' list, determine the important countries based on the following criteria:\\
    \hspace*{2em} a) Main Event: Countries directly related to the main event or topic of the article.\\
    \hspace*{2em} b) Role: Countries mentioned as main actors or playing a significant role in the article.\\
   \hspace*{1em} Provide the important countries as a comma-separated list of their 3-letter country codes (ISO 3166-1 alpha-3).\\
   
10. Verify if all important countries are correctly identified and listed based on the content provided. Answer `Yes' or `No'.\\

11. Explain why the important countries are not accurately identified if the answer to step 10 is `No'. If the answer to step 10 is `Yes', return `None'.\\

12. Provide the corrected list of important countries as a comma-separated list of 3-letter country codes (ISO 3166-1 alpha-3) if the answer to step 10 is `No'. If the answer to step 10 is `Yes', return `None'.\\

Article: \{text\}\\

Follow this format exactly to ensure proper parsing and then answer:\\
1. Countries: \{\{Answer here as a comma-separated list of 3-letter country codes, or "None".\}\}\\
2. Summary: \{\{Answer here with the summary text including all mentioned countries, up to 10 sentences. Ensure you include as much information from the original article as possible.\}\}\\
3. Countries Accurate: \{\{Answer `Yes' or `No'.\}\}\\
4. Explanation for Inaccuracy: \{\{If the answer to step 3 is `No', explain why the `Countries' list is not accurate. If the answer to step 3 is `Yes', return `None'.\}\}\\
5. Corrected Countries: \{\{If the answer to step 3 is `No', provide the corrected list of countries as a comma-separated list of 3-letter country codes (ISO 3166-1 alpha-3). If the answer to step 3 is `Yes', return `None'.\}\}\\
6. Summary Accurate: \{\{Answer `Yes' or `No'.\}\}\\
7. Explanation for Inaccuracy: \{\{If the answer to step 6 is `No', explain why the `Summary' is not accurate. If the answer to step 6 is `Yes', return `None'.\}\}\\
8. Corrected Summary: \{\{If the answer to step 6 is `No', provide the corrected summary, ensuring it includes all mentioned countries and follows the 10-sentence limit. If the answer to step 6 is `Yes', return `None'.\}\}\\
9. Important Countries: \{\{Answer here as a comma-separated list of 3-letter country codes based on the criteria.\}\}\\
10. Important Countries Accurate: \{\{Answer `Yes' or `No'.\}\}\\
11. Explanation for Inaccuracy: \{\{If the answer to step 10 is `No', explain why the important countries are not accurately identified. If the answer to step 10 is `Yes', return `None'.\}\}\\
12. Corrected Important Countries: \{\{If the answer to step 10 is `No', provide the corrected list of important countries as a comma-separated list of 3-letter country codes (ISO 3166-1 alpha-3) if the answer to step 10 is `Yes', return `None'.\}\}\\

        \end{minipage}
    }
    \caption{Extracting important countries scratchpad.}
    \label{app-fig:full-country-scratchpads}
\end{figure*}

The relationship labeling scratchpad (Figure \ref{appfig:label_pad}) focuses on identifying relationships between countries, summarizing the context, and correcting inaccuracies. This method allows for precise and comprehensive labeling of international relationships.

\begin{figure*}[h!]
    \centering
    \fbox{
        \begin{minipage}{1.0\textwidth}
        \fontsize{10}{12}\selectfont
You are provided with an article. Your tasks are: \\

1. Extract all countries mentioned in the article as a comma-separated list of their 3-letter country codes (ISO 3166-1 alpha-3). If no countries are mentioned, return "None".\\

2. Summarize the article in up to 10 sentences, ensuring all mentioned countries are included.\\

3. Verify if the extracted `Countries' list includes all countries mentioned in the article. Answer `Yes' or `No'.\\

4. Explain why the `Countries' list is not accurate if the answer to step 3 is `No'. If the answer to step 3 is `Yes', return `None'.\\

5. Provide the corrected list of countries as a comma-separated list of 3-letter country codes (ISO 3166-1 alpha-3) if the answer to step 3 is `No'. If the answer to step 3 is `Yes', return `None'.\\

6. Verify if the `Summary' of the article content is accurate and includes all mentioned countries. Answer `Yes' or `No'.\\

7. Explain why the `Summary' is not accurate if the answer to step 6 is `No'. If the answer to step 6 is `Yes', return `None'.\\

Article: \{text\}\\

Countries: \{country1\}, \{country2\} \\

Follow this format exactly to ensure proper parsing and then answer:\\
1. Countries: \{\{Answer here as a comma-separated list of 3-letter country codes, or "None".\}\}\\
2. Summary: \{\{Answer here with the summary text including all mentioned countries, up to 10 sentences. Ensure you include as much information from the original article as possible.\}\}\\
3. Countries Accurate: \{\{Answer `Yes' or `No'.\}\}\\
4. Explanation for Inaccuracy: \{\{If the answer to step 3 is `No', explain why the `Countries' list is not accurate. If the answer to step 3 is `Yes', return `None'.\}\}\\
5. Corrected Countries: \{\{If the answer to step 3 is `No', provide the corrected list of countries as a comma-separated list of 3-letter country codes (ISO 3166-1 alpha-3). If the answer to step 3 is `Yes', return `None'.\}\}\\
6. Summary Accurate: \{\{Answer `Yes' or `No'.\}\}\\
7. Explanation for Inaccuracy: \{\{If the answer to step 6 is `No', explain why the `Summary' is not accurate. If the answer to step 6 is `Yes', return `None'.\}\}\\

        \end{minipage}
    }
    \caption{Relationship labeling scratchpad.}

    \label{appfig:label_pad}

\end{figure*}

These structured prompts facilitate detailed and accurate annotations, critical for constructing a high-quality dataset for future event prediction.

\subsection{Logical Consistency in Scratchpads}
\begin{table}[]
\centering
\resizebox{\columnwidth}{!}{
\begin{tabular}{c|l|r}
\toprule
\textbf{Stage}                                                               & \multicolumn{1}{c|}{\textbf{Target Instruction}} & \multicolumn{1}{c}{\textbf{Consistency}} \\ \midrule
\multirow{5}{*}{\begin{tabular}[c]{@{}c@{}}Country\\ Extraction\end{tabular}} & Format Instruction (12)                    & 0.97                                      \\ \cmidrule{2-3} 
                                                                              & (S3=Yes, S4=None)                         & 1.00                                     \\ \cmidrule{2-3} 
                                                                              & (S3=Yes, S5=None)                         & 1.00                                    \\ \cmidrule{2-3} 
                                                                              & (S10=Yes, S11=None)                       & 1.00                                      \\ \cmidrule{2-3} 
                                                                              & (S10=Yes, S12=None)                       & 1.00                                      \\ \midrule
\multirow{4}{*}{\begin{tabular}[c]{@{}c@{}}Relationship\\ Labeling\end{tabular}} & Format Instruction (7)                     & 0.98                                      \\ \cmidrule{2-3} 
                                                                              & (S1=No, S{[}2:7{]}=None)                  & 1.00                                      \\ \cmidrule{2-3} 
                                                                              & (S4=Yes, S5=None)                         & 1.00                                      \\ \cmidrule{2-3} 
                                                                              & (S4=Yes, S6=None)                         & 1.00                                      \\ 
\bottomrule
\end{tabular}
}
\caption{Our two scratchpads include numerous conditional instructions and format guidelines for each instruction's outcome. Despite containing more instructions compared to existing annotation prompts, our scratchpad is efficiently managed by the latest LLM. There are very few logical inconsistencies in its responses.}
\label{tab:logical-consistency}
\end{table}
Annotation using LLMs has been widely adopted, showing increasingly successful results as LLM performance improves. Unlike existing works~\cite{tan2024large-survey,pmlr-v225-goel23a} that provide guidelines and a few examples for a single task, our scratchpads have more questions and strict format guidelines. Specifically, as shown in Figures \ref{app-fig:full-country-scratchpads} and \ref{appfig:label_pad}, our two scratchpads have a variety of conditional instructions and detailed format guidelines for each instruction's outcome. Despite this difficulty, we found that the LLM produces results consistent with the guidelines. Although they include many instructions and long prompts, the LLM manages them efficiently as demonstrated in Table~\ref{tab:logical-consistency}. Consequently, there are  few logical inconsistencies in its responses.

\subsection{Annotation Samples with Scratchpads}
\label{app-sec:scratchpad-samples}

The following figures illustrate examples of our self-correcting scratchpads used in the annotation process. These ensure the accurate extraction of important countries and the labeling of relationships between countries in news articles.

Figure \ref{fig:sample} demonstrates the process of self-correcting in extracting important countries from an article. It includes steps for identifying mentioned countries, verifying the accuracy of the list, and correcting it if necessary.

\begin{figure*}[h!]
    \centering
    \fbox{
        \begin{minipage}{1.0\textwidth}
        \fontsize{10}{12}\selectfont
\textbf{Article:} North Korea announced on Wednesday that it had launched its first reconnaissance satellite into orbit, claiming it as a defense measure against the dangerous military maneuvers of its enemies. The "Malligyong-1" satellite was launched on a new carrier rocket, the "Chollima-1," which North Korea stated was a legitimate right to strengthen its self-defense. However, South Korea, the United States, and Japan could not confirm if the satellite had entered orbit, with South Korea deeming it a violation of UN Security Council resolutions and ramping up military responses. Japan also condemned the launch, emphasizing cooperation with the US and South Korea. Experts suggested that a successful launch could significantly enhance North Korea's military capabilities, though there were doubts about its actual performance. North Korea announced further launches while criticizing the military actions of the US, South Korea, and Japan. Additionally, it remains uncertain if Russia aided North Korea, suggesting that North Korea might have resolved technical issues independently. Previously, North Korea experienced launch failures with rockets breaking into parts and falling into the Yellow Sea, East China Sea, and the Pacific Ocean. \\

\textbf{Step 1:} Extract all countries mentioned in the article as a comma-separated list of their 3-letter country codes (ISO 3166-1 alpha-3). If no countries are mentioned, return "None".\\
\textbf{Answer 1:} \underline{`RUS', `USA', `JPN', `CHN', `PRK', `KOR'} \\

\textbf{Step 2:} ... \\

\textbf{Step 3:} Verify if the extracted `Countries' list includes all countries mentioned in the article. Answer `Yes' or `No'.\\
\textbf{Answer 3:} \underline{No} \\

\textbf{Step 4:} Explain why the `Countries' list is not accurate if the answer to step 3 is `No'. If the answer to step 3 is `Yes', return `None'.\\
\textbf{Answer 4:} The provided list includes \uline{China (`CHN') which was mentioned only in the context of a sea name, not as an active country} in the article. \\

\textbf{Step 5:} Provide the corrected list of countries as a comma-separated list of 3-letter country codes (ISO 3166-1 alpha-3) if the answer to step 3 is `No'. If the answer to step 3 is `Yes', return `None'.\\
\textbf{Answer 5:} \underline{`RUS', `USA', `JPN', `PRK', `KOR'} \\

\textbf{Step 6:} ... \\

\textbf{Step 9:} From the extracted `Countries' list, determine the important countries based on the following criteria:
a) Main Event: Countries directly related to the main event or topic of the article.
b) Role: Countries mentioned as main actors or playing a significant role in the article.
Provide the important countries as a comma-separated list of their 3-letter country codes.\\
\textbf{Answer 9:} \underline{`RUS', `USA', `JPN', `PRK', `KOR'} \\

\textbf{Step 10:} Verify if all important countries are correctly identified and listed based on the content provided. Answer `Yes' or `No'.\\
\textbf{Answer 10:} \underline{No} \\

\textbf{Step 11:} Explain why the important countries are not accurately identified if the answer to step 10 is `No'. If the answer to step 10 is `Yes', return `None'.\\
\textbf{Answer 11:} The list of important countries includes \uline{Russia (`RUS'), which, although mentioned in the context of providing potential assistance, is not directly related to the main event} or playing a significant role in the article. \\

\textbf{Step 12:} Provide the corrected list of important countries as a comma-separated list of 3-letter country codes (ISO 3166-1 alpha-3) if the answer to step 10 is `No'. If the answer to step 10 is `Yes', return `None'.\\
\textbf{Answer 12:} \underline{`PRK', `KOR', `USA', `JPN'} \\

        \end{minipage}
    }
    \caption{Example of self-correcting in extracting important countries in an article.}
    \label{fig:sample}

\end{figure*}

Figures \ref{fig:sampletext} and \ref{fig:sampletextttt} show examples of self-correcting in labeling the relationship between two countries. These examples include steps for scoring the relationship, providing explanations, verifying the accuracy of the score, and correcting it if needed.

\begin{figure*}[h!]
    \centering
    \fbox{
        \begin{minipage}{1.0\textwidth}
        \fontsize{10}{12}\selectfont
\textbf{Article:} Europe's slow vaccination program has given Russia an opportunity for diplomatic gain. A video conference between French President Macron, German Chancellor Merkel, and Russian President Putin discussed the potential registration and production of the Russian Sputnik V vaccine in the EU. Some EU member states and allies fear Russia's intentions, seeing the vaccine as a means to divide the continent. Hungary and Slovakia have already ordered Sputnik doses, while Austria, Italy, and others are in talks for production and procurement. Lithuania’s Prime Minister openly criticized Putin's motives, contrasting the relaxed stance of many EU members. Some fear Russia’s vaccine as a diplomatic tool and a potential source of division within the EU. The EU’s centralized vaccine procurement faces criticism, and the notion of using Sputnik draws mixed reactions. Russian analysts suggest that Russia aims to win a soft power battle by getting the vaccine recognized in Europe. EU internal divisions over the vaccine reflect broader diplomatic and geopolitical tensions. Allegations of hacking and broader concerns over Russia's intentions add complexity to the situation. \\

\textbf{Countries:} {`FRA', `RUS'} \\

\textbf{Step 1:} Determine if there is evidence to predict the relationship between `FRA' and `RUS' as either `Conflict', `Cooperation', or `Neutral' based on the full text of the article. Note that simply categorizing it as `Neutral' because it is neither `Conflict' nor `Cooperation' does not qualify as evidence. Answer with `Yes' or `No'.\\
\textbf{Answer 1:} \underline{Yes} \\

\textbf{Step 2:} Score the relationship based on the evidence on a scale from 0.0 to 1.0, where 0.0 signifies complete `Cooperation' and 1.0 signifies complete `Conflict'; provide only the numeric value (e.g., 0.8). Return `None' if the answer to Step 1 is `No'.\\
\textbf{Answer 2:} \underline{Conflict with 0.7 score} \\

\textbf{Step 3:} Explain your score with evidence from the article. Return `None' if the answer to Step 1 is `No'.\\
\textbf{Answer 3:} The article discusses the use of the Russian Sputnik V vaccine as diplomatically divisive. It highlights that the discussion between Macron and Putin over the vaccine has caused significant concern among many EU leaders and diplomats who view it as a tool being potentially used by Russia to exploit divisions within Europe. While there is some cooperative dialogue regarding the vaccine, \uline{the overall tone and concern about Russia's intentions and the potential for it to deepen divisions suggest a strained relationship indicative of conflict rather than cooperation.} \\

\textbf{Step 4:} Verify if your score accurately reflects the full text of the article. Answer with `Yes' or `No'. Return `None' if the answer to Step 1 is `No'.\\
\textbf{Answer 4:} \underline{No} \\

\textbf{Step 5:} Explain why your score does not accurately reflect the full text of the article. Return `None' if the answer to Step 1 is `No' or if the answer to Step 4 is `Yes'.\\
\textbf{Answer 5:} Although there is a significant element of distrust and concern about Russia's intentions, there are also instances of cooperation and willingness to engage from France's perspective. \uline{The initial score might have overemphasized the conflict aspects.} \\

\textbf{Step 6:} Provide a corrected score that you believe accurately reflects the full text. Return `None' if the answer to Step 1 is `No' or if the answer to Step 4 is `Yes'.\\
\textbf{Answer 6:} \underline{Neutral with 0.5 score} \\

\textbf{Step 7:} Explain why you gave the corrected score. Return `None' if the answer to Step 1 is `No' or if the answer to Step 4 is `Yes'.\\
\textbf{Answer 7:} A score of 0.5 better reflects the nuanced nature of the relationship, which involves both cooperation (in terms of vaccine discussion) and conflict (due to the underlying distrust and geopolitical maneuvering). \uline{It captures the balance between cooperative engagement on vaccination and the significant concerns about Russia's strategic intentions.}

        \end{minipage}
    }
    \caption{Example of self-correcting in labeling relationship between two countries.}
    \label{fig:sampletext}
\end{figure*}

\begin{figure*}[h!]
    \centering
    \fbox{
        \begin{minipage}{1.0\textwidth}
        \fontsize{10}{12}\selectfont
\textbf{Article:} The United States, South Korea, and Japan announced renewed commitments at a summit at Camp David to show solidarity against China. The leaders launched new military exercises, crisis communication hotlines, and yearly trilateral summits. President Biden emphasized the summit's focus on deepening Indo-Pacific alliances amid concerns over China's behavior, North Korea's provocations, and Russia's war in Ukraine. Chinese officials criticized the summit as a threat to regional stability. Despite historical tensions between Japan and South Korea, their leaders have worked to mend relations in response to common threats. Biden reinforced his commitment to maintaining and strengthening alliances in Asia, including pledges for international law and stability in the Taiwan Straits and South China Sea. The joint statement condemned North Korea's missile programs and illicit cyber activities. The summit marked Biden's first hosting of foreign leaders at Camp David, underscoring his administration's focus on reinvigorating alliances. The leaders' cooperation aims to advance regional and global security and prosperity. The trilateral partnership is seen as crucial for responding to geopolitical challenges. \\

\textbf{Countries:} {`CHN', `PRK'} \\

\textbf{Step 1:} Determine if there is evidence to predict the relationship between `FRA' and `RUS' as either `Conflict', `Cooperation', or `Neutral' based on the full text of the article. Note that simply categorizing it as `Neutral' because it is neither `Conflict' nor `Cooperation' does not qualify as evidence. Answer with `Yes' or `No'.\\
\textbf{Answer 1:} \underline{Yes} \\

\textbf{Step 2:} Score the relationship based on the evidence on a scale from 0.0 to 1.0, where 0.0 signifies complete `Cooperation' and 1.0 signifies complete `Conflict'; provide only the numeric value (e.g., 0.8). Return `None' if the answer to Step 1 is `No'.\\
\textbf{Answer 2:} \underline{Conflict with 0.7 score} \\

\textbf{Step 3:} Explain your score with evidence from the article. Return `None' if the answer to Step 1 is `No'.\\
\textbf{Answer 3:} The article discusses increased tensions and animosity between the US and China due to renewed commitments and military exercises among the US, Japan, and South Korea. It also includes China's unhappy reaction to these developments. The article mentions North Korea's controversial actions, including its missile program and potential arms transfers to Russia. These points indicate potential conflict or strain in the relationship between China (CHN) and North Korea (PRK) since increased tensions with the US might add strain to China's relations with its allies, including North Korea.\\

\textbf{Step 4:} Verify if your score accurately reflects the full text of the article. Answer with `Yes' or `No'. Return `None' if the answer to Step 1 is `No'.\\
\textbf{Answer 4:} \underline{No} \\

\textbf{Step 5:} Explain why your score does not accurately reflect the full text of the article. Return `None' if the answer to Step 1 is `No' or if the answer to Step 4 is `Yes'.\\
\textbf{Answer 5:} The score might be too high, indicating a stronger conflict than suggested by the text. The specific conflict between China and North Korea isn't deeply elaborated upon, thus a somewhat lower level of conflict might be more accurate. \\

\textbf{Step 6:} Provide a corrected score that you believe accurately reflects the full text. Return `None' if the answer to Step 1 is `No' or if the answer to Step 4 is `Yes'.\\
\textbf{Answer 6:} \underline{Neutral with 0.5 score} \\

\textbf{Step 7:} Explain why you gave the corrected score. Return `None' if the answer to Step 1 is `No' or if the answer to Step 4 is `Yes'.\\
\textbf{Answer 7:} The relationship might be more strained than fully conflicting, as there are shared interests and historical ties between China and North Korea that might prevent full-scale conflict even amid increasing tensions related to the US-led alliance activities.

        \end{minipage}
    }
    \caption{Example of self-correcting in labeling relationship between two countries.}
    \label{fig:sampletextttt}
\end{figure*}

These samples highlight the effectiveness of our self-correcting mechanism in improving the accuracy and reliability of our dataset annotations.

\subsection{Detailed Dataset Construction}
To collect comprehensive information describing international political events, we need to create a dataset that includes reliable, high-quality news articles and detailed annotations of international relations. Our dataset construction process involves three main steps: Data Collection, Subject Extraction, and Score Labeling. We have automated this entire process and the dataset will be publicly available. Additionally, we will release the code used for this automated process. By making this code publicly available, we aim to support the analysis of international relations and various other text-based event analysis tasks across different domains.

\subsubsection{Challenges and Requirements}
Creating a high-quality dataset for international political events presents several key challenges:
\begin{enumerate}
    \item \textbf{Capturing Multiple Subjects}: Ensuring that all relevant countries and their relationships are captured in each event. A single news article or event often involves multiple countries with complex interactions. It is essential to identify and label all significant subjects to provide a comprehensive understanding of the event dynamics.
    \item \textbf{Scoring Relationship Labels}: Representing relationships with numerical scores to capture the nuance of interactions. For instance, simply labeling relationships as "good" or "bad" can miss the subtleties of international dynamics. Numerical scores allow for a more nuanced representation, reflecting varying degrees of cooperation or conflict.
    \item \textbf{Handling Unknown Relationships}: Identifying and labeling relationships that cannot be determined. In many cases, the information available might not be sufficient to ascertain the nature of the relationship between countries. It is crucial to accurately label these instances as "unknown" to avoid misleading conclusions.
    \item \textbf{Ensuring Consistency and Reliability}: Achieving consistency and reliability in the information extracted from articles. Variations in labeling due to different interpretations or extraction errors can lead to unreliable models. Techniques such as ensemble labeling and the use of self-correcting prompts help ensure that the extracted information is accurate and consistent across different sources and iterations.
\end{enumerate}

To address these challenges, we employ a systematic approach involving advanced techniques in data curation, subject extraction, and score labeling.

\subsubsection{Data Curation}
The first step is to gather relevant news articles that cover a wide temporal range and include diverse perspectives on international events. We use the GDELT project primarily for acquiring news links due to its comprehensive coverage of global events. Our dataset covers news articles from February 2013 to May 2024, ensuring a wide temporal range for comprehensive analysis.

\paragraph{Keyword Filtering}
Given the vast number of news articles available, filtering them to ensure relevance is crucial. We employed an extensive list of keywords related to international relations to capture as many relevant articles as possible, retaining documents where any of these keywords appeared at least once. The complete list of keywords are as follows: [`Diplomacy', `Trade', `Military', `Sanctions', `United Nations', `NATO', `G7', `G20', `Security', `Foreign', `Territorial', `Rights', `Conference', `International', `Law', `Peace', `Cooperation', `Border', `Visa', 'Immigration', `Refugee', `Terrorism', `Nuclear', `Climate', `Dispute', `Maritime', `Cybersecurity', `Global', `Economy', `Humanitarian', `Aid', `War', `Defense', `Court', `Conflict', `Embassy', `Budget', `Envoy', `Mediation', `Resolution', `Finance', `Development', `Change', `Crisis', `Relief', `Control', `Regional', `Alliance', `Negotiation', `Peacekeeping', `Multilateral']

\paragraph{Selecting Publisher}
Even after keyword filtering, many articles report on the same events, leading to content duplication. To address this, we limit our sources to a single reliable publisher. This strategy helps eliminate duplicate articles about the same event and avoids including news from sources with questionable reliability. We selected a large and reputable publisher, such as CNN, known for its comprehensive coverage and data trustworthiness.

\subsubsection{Content Extraction}
Once we have curated the relevant news articles, the next step is to extract meaningful information related to international relations. The main issues here are accurate extraction and the need to handle diverse and nuanced information.

\paragraph{All Subjects Extraction}
We start by extracting all countries mentioned in each news article as subjects, ensuring that any country mentioned even once is included. To ensure uniformity, we represent each country using its three-letter code according to the ISO 3166-1 alpha-3 standard (e.g., USA, RUS, CHN).

\paragraph{Extracting Key Subjects}
Not all extracted subjects play a significant role in the events described in the articles. For example, a mention of the USA in the context of "a person born in the USA" might not be relevant to the international event being reported. Therefore, within the same prompt framework, we identify and retain only the key subjects central to the events described in the news articles. This step helps us focus on the most relevant entities involved in the news.

\paragraph{Generating Summaries}
News articles can vary significantly in length, and excessively long articles can pose challenges for many NLP models. To enhance usability and maintain core information, we summarize each article to no more than 10 sentences. This summary captures the essence of the article while keeping the length within roughly 512 tokens, optimizing it for subsequent processing. Note that we will release both the original articles and their corresponding detailed summaries, ensuring full transparency and utility for various applications.

\paragraph{Scratchpad Design}
We employ a scratchpad design to structure and refine the extracted content. The scratchpad methodology allows us to systematically capture and organize relevant information from the articles. We designed the scratchpad to check and correct initial outputs, enhancing the reliability of subject extraction and summarization. These features are challenging to aggregate through averaging or voting; hence, the scratchpad is particularly effective in this situation. This also allows us to interpret the intent of the output more accurately. 

\subsubsection{Label Annotation}
Next, we create pairs of all key subjects identified in each article and label their relationships based on the original text. This addresses the challenge of scoring relationship labels by providing a nuanced representation of the interactions.

\paragraph{Scoring Assignment}
We use LLMs to assign scores to the relationships between the key subjects. For example, if there are four key subjects, we generate six pairs. It is essential to express the relationships beyond just good or bad, so each pair is assigned a score between 0 and 1, where values closer to 1 indicate conflict and values closer to 0 indicate cooperation. Scores around 0.5 suggest a indeterminate relationship. This scoring helps quantify the nature of relationships for more precise predictions.

\paragraph{Ensemble Labeling}
To improve the reliability of our numerical score labels, we utilize an ensemble labeling approach. We generate five sets of annotations through distinct runs using GPT-4. If more than half of the runs label a relationship as unknown, we assign the unknown label to that pair. Otherwise, we average the scores from the known labels to determine the final score. This method enhances the robustness and accuracy of our labeling process, and can be considered a form of consistency check.

Through these carefully designed stages, we have developed a dataset that accurately and comprehensively reflects the intricate dynamics of international relations. This collection establishes a new standard for text-based future event prediction, ensuring a high level of detail, reliability, and usability.

\paragraph{Time Efficiency Analysis}
We conducted additional experiments to analyze the time efficiency of the annotation process, specifically for the country extraction and relationship labeling stages, with and without the self-correcting mechanism.

For the country extraction stage, which involves 12 instructions, the process takes approximately 10.19 seconds on average. When the self-correcting steps are removed, the number of instructions reduces to 2, and the process takes around 8.85 seconds. The average query length for the self-correcting version is 8270.65 tokens, while the version without self-correcting has an average length of 5038.65 tokens.

For the relationship labeling stage, the process takes an average of 10.85 seconds to process 7 instructions, while removing the self-correcting mechanism reduces the instructions to 2, resulting in a response time of 9.97 seconds. The average query length for the self-correcting version is 6952.92 tokens, compared to 5620.62 tokens for the non-self-correcting version.

Although the query length and the number of steps increase due to the self-correcting mechanism, the overall time required does not significantly increase. This indicates that the self-correcting mechanism improves the accuracy of the annotations without a substantial impact on time efficiency.

\section{Experiment Details}

\subsection{Settings and Hyperparameters for Sentence Classification Model}
In this section, we provide detailed settings and hyperparameters used in our document classification experiments. The goal of these experiments is to evaluate the reliability of our annotated labels by comparing models trained on our dataset with those trained on the GDELT dataset.

\subsubsection{Environment and Setup}
All experiments were conducted using PyTorch and the Hugging Face Transformers library. The training and evaluation were performed on a machine equipped with NVIDIA GPUs to leverage hardware acceleration.

\subsubsection{Data Preparation}
The data used for training and evaluation was sourced from two main datasets: our annotated dataset and the GDELT dataset. For fair comparison, we selected samples where both our labels and domain expert labels were not classified as `Unknown'. This resulted in a two-class classification task (`Cooperation' vs. `Conflict').

\begin{itemize}
\item \textbf{Training Data:} A subset of the data was used for training the models. The train dataset was divided into training and validation sets to monitor the model performance during training.
\item \textbf{Test Data:} The test dataset was labeled by domain experts and was used for evaluating the model performance.
\end{itemize}

\subsubsection{Model Selection}
We evaluated several pre-trained language models known for their effectiveness in text classification tasks:
\begin{itemize}
\item BERT \citep{devlin2018bert}
\item RoBERTa \citep{liu2019roberta}
\item ALBERT \citep{lan2019albert}
\item XLNet \citep{yang2019xlnet}
\item DistilBERT \citep{sanh2019distilbert}
\end{itemize}

\subsubsection{Hyperparameters}
The following hyperparameters were used for training the models:

\begin{itemize}
\item \textbf{Epochs:} 20
\item \textbf{Batch Size:} 16
\item \textbf{Learning Rate:} 2e-5
\item \textbf{Max Sequence Length:} 512 tokens
\item \textbf{Optimizer:} AdamW
\item \textbf{Scheduler:} Linear scheduler with warm-up steps
\end{itemize}

\subsubsection{Training Procedure}
\begin{enumerate}
\item \textbf{Data Loading:} The data was tokenized using the respective tokenizer for each pre-trained model.
\item \textbf{Model Training:} Models were trained using the training dataset with the specified hyperparameters. Gradient accumulation and mixed-precision training were used to optimize memory usage and speed up training.
\item \textbf{Validation:} During training, model performance was monitored using the validation dataset. The model with the best validation loss was saved.
\item \textbf{Evaluation:} The best-performing model on the validation set was evaluated on the test set. The performance metrics included accuracy, F1 score, and confusion matrix.
\end{enumerate}

\section{Information for Data Usage Compliance}
\subsection{Citation of Artifact Creators}
We utilized the Global Database of Events, Language, and Tone (GDELT) dataset in our research. The GDELT Project, supported by Google Jigsaw, monitors global news and provides a comprehensive archive for research and analysis. We used the latest version of the GDELT dataset, and we hereby cite the original creators:
\begin{itemize}
    \item Leetaru, Kalev, and Philip A. Schrodt. "GDELT: Global Data on Events, Location and Tone, 1979-2012." ISA Annual Convention (2013). Available at: https://www.gdeltproject.org/
\end{itemize}

\subsection{License and Terms of Use}
The GDELT Project's datasets are released under terms that allow for unlimited and unrestricted use for academic, commercial, or governmental purposes without a fee. Redistribution, rehosting, republishing, and mirroring of the GDELT datasets in any form is permitted, provided that any use or redistribution includes a citation to the GDELT Project and a link to the website (https://www.gdeltproject.org/).

\subsection{Consistency with Intended Use}
Our use of the GDELT dataset is consistent with its intended use as outlined by the GDELT Project. The dataset is intended for research and analysis of global society, and our research falls under this category. We used the data to enhance our understanding of international relations and predict future geopolitical events, which aligns with the GDELT Project's goal of enabling research on human societal behavior and beliefs.

\subsection{Data Anonymization}
The GDELT dataset consists of publicly available news articles and does not contain personally identifiable information. The data is aggregated and focuses on events, locations, and entities rather than individuals.

\subsection{Use of AI Assistants}
We used AI assistants in our research, coding, and writing processes. Specifically, we employed ChatGPT for refining writing, and aiding in the coding tasks. The assistance from AI tools was instrumental in enhancing productivity and ensuring clarity in our documentation and experimental design. The use of these tools was conducted in accordance with ethical guidelines and with careful oversight to maintain the integrity and originality of our research. 

\end{document}